\documentclass{article}
\usepackage{spconf,amsmath,graphicx,hyperref}
\usepackage{multirow}
\usepackage{bbding}
\usepackage{booktabs}
\usepackage{graphicx}
\usepackage{amsmath}
\usepackage{pgfplots}

\definecolor{mygreen}{HTML}{008000}

\title{Chain-of-Thought Re-ranking for Image Retrieval Tasks}
\name{Author(s) Name(s)}
\address{Author Affiliation(s)}
\name{Shangrong Wu$^{\star}$ \qquad Yanghong Zhou$^{\dagger}$ \qquad Yang Chen$^{\dagger}$ \qquad Feng Zhang$^{\dagger}$ \qquad P.~Y.~Mok$^{\ddagger}$\thanks{P. Y. Mok is the corresponding author}}
  
  \address{$^{\star}$ National University of Singapore, Singapore. e1504074@u.nus.edu\\
      $^{\dagger}$ The Hong Kong Polytechnic University, Hong Kong. \\ \{yanghong2.zhou, ethan.zhang\}@polyu.edu.hk, chenyang001001@gmail.com \\
      $^\ddagger$ The Hong Kong University of Science and Technology, Hong Kong. tracy.mok@ust.hk}

\begin{document}
\ninept
\maketitle
\begin{abstract}
Image retrieval remains a fundamental yet challenging problem in computer vision. While recent advances in Multimodal Large Language Models (MLLMs) have demonstrated strong reasoning capabilities, existing methods typically employ them only for evaluation, without involving them directly in the ranking process. As a result, their rich multimodal reasoning abilities remain underutilized, leading to suboptimal performance.
In this paper, we propose a novel Chain-of-Thought Re-Ranking (CoTRR) method to address this issue. Specifically, we design a listwise ranking prompt that enables MLLM to directly participate in re-ranking candidate images. This ranking process is grounded in an image evaluation prompt, which assesses how well each candidate aligns with user's query. By allowing MLLM to perform listwise reasoning, our method supports global comparison, consistent reasoning, and interpretable decision-making—all of which are essential for accurate image retrieval. To enable structured and fine-grained analysis, we further introduce a query deconstruction prompt, which breaks down the original query into multiple semantic components.
Extensive experiments on five datasets demonstrate the effectiveness of our CoTRR method, which achieves state-of-the-art performance across three image retrieval tasks, including text-to-image retrieval (TIR), composed image retrieval (CIR) and chat-based image retrieval (Chat-IR) . Our code is available at \url{https://github.com/freshfish15/CoTRR}.

\end{abstract}
\begin{keywords}
Text-to-Image retrieval, composed image retrieval, chain-of-thought reasoning,  large  models. 
\end{keywords}
\section{Introduction}
\label{sec:intro}
Image retrieval has been a fundamental research problem in computer vision since the 1970s~\cite{datta2008image}, and it continues to play a crucial role in  applications such as visual search, search engines, and e-commerce platforms.
However, accurately and efficiently retrieving results that match a user's input remains challenging, due to the inherent ambiguity and limitations in interpreting user input, as well as the difficulty in aligning the input with the target content—especially when the modalities differ, as in tasks like TIR and CIR. To address these challenges, some previous works \cite{Dialog-based,chatir} have explored Chat-IR systems, which aim to better understand user intent through interactive dialogue. Meanwhile, some other approaches \cite{ma2025multi, wen2023target} focus on enhancing the accuracy and efficiency of input-target matching by learning generalized and robust representations, but they typically require additional training or fine-tuning. 

Recently, owing to the rapid development of large language models (LLMs) and visual language models (MLLMs) and their powerful capabilities in language understanding, reasoning, and processing multimodal information, some works~\cite{chatir,plugir,osrcir, cotmr,imagescope} have leveraged LLMs for image retrieval without requiring additional training.
For CIR, OSrCIR \cite{osrcir} and CoTMR \cite{cotmr} utilize large language models to infer the user's intention through joint analysis of the image and textual query.
By leveraging Chain-of-Thought (CoT) reasoning~\cite{wei2022chain,zhang2023multimodal} to enhance the analysis capabilities of LLM, these methods significantly improve the accuracy and efficiency of training-free CIR.
However, these methods are limited to either chat-based image retrieval or composed image retrieval.
To address this limitation,  ImageScope \cite{imagescope} proposes a unified framework for TIR, CIR and Chat-IR tasks. 
More specifically, it first employs a vision-language model (VLM) to convert all three tasks into a standardized text-to-image retrieval process, and then leverages a large language model (LLM) to verify and evaluate the retrieved results.
However, these methods typically use MLLMs only for evaluating candidate images after retrieval, without directly involving them in the re-ranking process.
As a consequence, the rich multimodal reasoning capabilities of MLLMs remain largely underutilized, resulting in limited global comparison and a fragmented decision-making process during re-ranking.


\begin{figure*}[t]
    \centering
    \includegraphics[width=\linewidth]{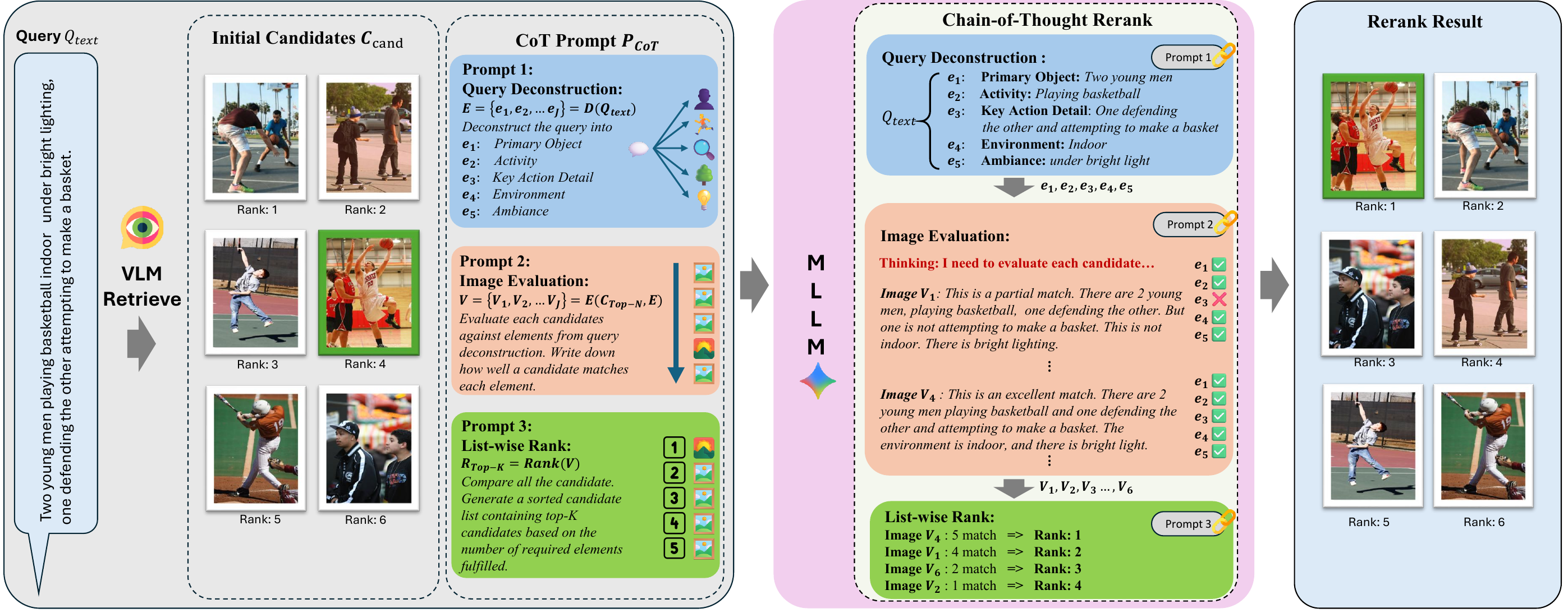}
    \caption{The overall architecture of our CoTRR.}
    \label{fig:framework-overview}
\end{figure*}

In this paper, we propose a novel chain-of-thought re-ranking (CoTRR) method to enhance the ranking of target images, which can be easily and seamlessly applied to text-to-image retrieval, composed image retrieval, and chat-based image retrieval. To this end, the initial retrieval results are first generated using a retrieval method, and then CoTRR is employed to re-rank the top-K candidates in one stage.
More specifically, CoTRR introduces a query deconstruction prompt to thoroughly understand the user's intention and to reformat the input into a standardized structure, enabling more effective comparison between the user’s input and candidate images.
Additionally,  an image evaluation paradigm is proposed to evaluate how well the candidates match with the input. 
Unlike   ImageScope \cite{imagescope} which evaluates whether a candidate meets multi-granularity requirements with a simple yes-or-no judgment, our approach is more intuitive and offers richer, more informative insights for ranking analysis.
Finally, based on the evaluation results, a listwise ranking prompt is employed to compare these evaluations and generate a new ranking order for the candidate images. 
By directly leveraging the MLLM for ranking based on its evaluation, our approach preserves consistent reasoning and facilitates globally informed ranking decisions, which are critical for improving retrieval performance.
It is also worth noting that, since our CoTRR does not depend on any specific retrieval method for the initial retrieval results, it can be applied to a wide range of retrieval tasks.
The main contributions of this work are as follows:
\begin{itemize}

    \item We propose {CoTRR}, a novel chain-of-thought re-ranking method that enhances the ranking of target images and is compatible with various retrieval tasks.
    \item CoTRR introduces a {query deconstruction prompt} and a {listwise ranking prompt}, along with an image evaluation paradigm, to better understand user intention and to more effectively evaluate, analyze, and re-rank the retrieved results.
    \item Experimental results on five datasets demonstrate that our CoTRR achieves state-of-the-art performance on three retrieval tasks. Ablation studies and thorough analyses provide additional evidence of CoTRR’s effectiveness.
    

    \end{itemize}

\section{Methodology}

Figure~\ref{fig:framework-overview} presents the overall architecture of the proposed \textbf{CoTRR} framework. The input query $Q$ can take two forms, depending on the retrieval task: (1) a text query $Q = Q_{\text{text}}$, used in TIR and Chat-IR or (2) a composed input $Q = \{I_r, T_m\}$, where $I_r$ is a reference image and $T_m$ is a manipulation text specifying the intended semantic modification of the reference image, commonly used in CIR.
Given the query, a retrieval model is first employed to obtain the top-K candidate images, denoted as $\mathcal{C}_{\text{cand}} = \{c_1, c_2, \ldots, c_K\}$. 
For TIR and Chat-IR  tasks, we use a Contrastive Language-Image Pre-training (CLIP) model \cite{clip1} to extract aligned text and image features, and compute their cosine similarity to retrieve the top-K candidate images.
For the composed image retrieval task, in order to better capture the user's intended semantic modification, we adopt OSrCIR \cite{osrcir} framework to obtain the initial candidate images. For Chat-IR, we follow the approach of PlugIR \cite{plugir} to formulate a text query based on dialogue history to obtain the initial candidates.
These initial candidates are then passed to the proposed CoTRR , which re-ranks them to produce a more accurate ranking result.
The CoTRR  consists of three key components: {query deconstruction}, {image evaluation}, and {listwise ranking}. Each component is described in detail in the following sections.

\vspace{-10pt}
\begin{table*}[h]
\caption{Comparison results for CIR task on CIRR and CIRCO. We implement OSrClR with OpenCLIP as backbone for fair comparsion.}
\label{tab:CIRResult}
\centering
\scriptsize
\begin{tabular}{llcccccc|ccccc}
\hline
\multicolumn{1}{c}{\multirow{3}{*}{Backbone/VLM}} & \multicolumn{1}{c}{\multirow{3}{*}{Method}} & \multicolumn{6}{c|}{CIRR} & \multicolumn{4}{c}{CIRCO} \\ \cline{3-12} 
\multicolumn{1}{c}{} & \multicolumn{1}{c}{} & \multicolumn{3}{c}{R@k} & \multicolumn{3}{c|}{$\text{R}_{\text{Subs}}\text{@k}$} & \multicolumn{4}{c}{mAP@k} \\ \cline{3-12} 
\multicolumn{1}{c}{} & \multicolumn{1}{c}{} & k=1 & k=5 & k=10 & k=1 & k=2 & k=3 & k=5 & k=10 & k=25 & k=50 \\ \hline
\multicolumn{1}{c}{\multirow{4}{*}{CLIP-ViT-B/32}}
& CIReVL \cite{CIReVL} & 23.94 & 52.51 & 66.00 & 60.17 & 80.05 & 90.19 & 14.94 & 15.42 & 17.00 & 17.82 \\
\multicolumn{1}{c}{} & OSrCIR \cite{osrcir} & 25.42 & 54.54 & 68.19 & 62.31 & 80.86 & 91.13 & 18.04 & 19.17 & 20.94 & 21.85 \\
\multicolumn{1}{c}{} & OSrCIR\cite{osrcir} (our impl)  & 32.63 & 61.23 & 74.02 & 64.34 & 82.07 & 91.69 & 20.56 & 21.11 & 23.07 & 24.01\\
\multicolumn{1}{c}{} & ImageScope \cite{imagescope} & 38.43 & 66.27 & 76.96 & 75.93 & 89.21 & 94.63 & 25.26 & 25.82 & 27.15 & 28.11 \\
\multicolumn{1}{c}{} & CoTRR & \textbf{50.84} & \textbf{72.80} & \textbf{77.83} & \textbf{83.11} & \textbf{93.11} & \textbf{97.33} & \textbf{41.36} & \textbf{41.31} & \textbf{42.61} & \textbf{42.98} \\ \hline
\multicolumn{1}{c}{\multirow{4}{*}{CLIP-ViT-L/14}} & CIReVL \cite{CIReVL} & 24.55 & 52.31 & 64.92 & 59.54 & 79.88 & 89.69 & 18.57 & 19.01 & 20.89 & 21.80 \\
\multicolumn{1}{c}{} & OSrCIR \cite{osrcir} & 29.45 & 57.68 & 69.86 & 62.12 & 81.92 & 91.10 & 23.87 & 25.33 & 27.84 & 28.97 \\
\multicolumn{1}{c}{} & OSrCIR (our impl) \cite{osrcir} & 33.49 & 63.33 & 75.06 & 64.34 & 82.07 & 91.67 & 25.19 & 26.21 & 28.39 & 29.38 \\
\multicolumn{1}{c}{} & ImageScope \cite{imagescope} & 39.37 & 67.54 & 78.05 & 76.36 & 89.40 & 95.21 & 28.36 & 29.23 & 30.81 & 31.88 \\
\multicolumn{1}{c}{} & CoTRR & \textbf{50.48} & \textbf{73.69} & \textbf{79.06} & \textbf{83.45} & \textbf{93.25} & \textbf{97.16} & \textbf{45.05}  &\textbf{45.22}  &\textbf{46.90}  &\textbf{47.36}
\\
\hline
\end{tabular}
\end{table*}

\begin{table*}[h]
    \centering
    \caption{Comparsion Results for TIR task on Flickr30K and MSCOCO datasets.}
    \label{tab:TIRResults}
    \setlength{\tabcolsep}{6pt} 
    \begin{tabular}{l|ccc|ccc|ccc}
        \toprule
        \multirow{2}{*}{Method} & \multicolumn{3}{c}{Flickr30K} & \multicolumn{3}{c}{MSCOCO} & \multicolumn{3}{c}{Average} \\
        & R@1 & R@5 & R@10 & R@1 & R@5 & R@10 & R@1 & R@5 & R@10 \\
        \midrule
        CLIP-ViT-B/32  & 66.56 & 88.16 & 93.02 & 39.45 & 65.51 & 75.65 & 53.01 & 76.83 & 84.33 \\
        ImageScope \cite{imagescope}  & 78.84 & 92.66 & 95.64 & 51.23 & 73.32 & 80.79 & 65.04 & 82.99 & 88.22 \\
        CoTRR   & \textbf{84.68} & \textbf{94.72} & \textbf{96.08} & \textbf{58.31} & \textbf{77.23} & \textbf{81.77} & \textbf{71.50} & \textbf{85.97} & \textbf{88.92} \\
        \midrule
        CLIP-ViT-L/14 & 75.72 & 92.96 & 96.00 & 46.46 & 71.10 & 79.78 & 61.09 & 82.03 & 87.89 \\
        ImageScope \cite{imagescope}  & 81.10 & 94.02 & 96.82 & 53.73 & 75.96 & 83.50 & 67.42 & 84.99 & 90.16 \\
        CoTRR   & \textbf{86.83} & \textbf{96.49} & \textbf{97.57} & \textbf{59.92} & \textbf{79.77} & \textbf{84.71} & \textbf{73.37} & \textbf{88.13} & \textbf{91.14} \\
        \bottomrule
    \end{tabular}
\end{table*}

\subsection{Query Deconstruction}

Query deconstruction is designed to transform unstructured textual descriptions into structured semantic representations, enabling more accurate and interpretable downstream evaluation and reasoning.

For TIR and Chat-IR, the input query is $Q_{\text{text}}$ while CIR takes the manipulation text $T_m$  and the reference image $I_{r}$ as the input.
Based on linguistic and contextual cues, each query is decomposed into five core semantic components: primary subject (e.g., two young men), activity (e.g., playing basketball), key details (e.g., one defending the other and attempting to make a basket), environment (e.g., indoor), and ambiance (e.g., under bright light),  denoted as $E=\{e_i\}_{i=1}^M$. 
This process can be denoted as:
\begin{equation}
E = 
\begin{cases}
\text{Deconstruct}(Q_{\text{text}}), & \text{for TIR and Chat-IR} \\
\text{Deconstruct}(T_{m}, I_{r}), & \text{for CIR}
\end{cases}
\end{equation}
This structured representation facilitates a more consistent and interpretable comparison with candidate images during the evaluation and re-ranking stages.

\subsection{Image evaluation}
The goal of the image evaluation stage is to assess whether each retrieved candidate image satisfies the user's input, specifically the deconstructed query. To avoid overly stringent or binary judgments that may lead to inaccurate evaluations, we prompt the language model to provide a detailed explanation of how well each candidate image aligns with each individual semantic component $e_i$ of the deconstructed query, rather than directly requesting a simple "yes" or "no" response as in ImageScope. Specifically, the model is instructed to assign an overall qualitative judgment (e.g., "partial match", "excellent match") for the entire candidate image, followed by a concise rationale that explains how well the image aligns with each individual semantic component. Two representative examples are shown in Figure~\ref{fig:framework-overview}.
Since the feedback provided by this evaluation method is both structured and fine-grained, it supplies the subsequent re-ranking module with richer and more accurate information, thereby facilitating enhancing ranking performance. 
Formally, given a candidate image $c \in \mathcal{C}_{\text{cand}}$ and the deconstructed query $E = \{e_i\}_{i=1}^5$, the evaluation output can be defined as:
\begin{equation}
    v = \text{Evaluate}(c, E)
\end{equation}
where $v$ denotes the textual evaluation result produced by the MLLM, which includes both the qualitative overall judgment and component-wise explanation.

\subsection{List-wise ranking}
With the evaluation results of the top-K candidate images, represented as $\mathcal{V} = \{v_1, v_2, \ldots, v_K\}$, we prompt the language model to review the textual evaluation $v_i$ for each candidate and perform a comparative analysis across all candidates.  
Based on this comparison, a sorted candidate list $\mathcal{R}_{\text{Top-K}}$ containing the top-$K$ images is generated. For example, candidate image $I_4$ fulfills all deconstructed components and is thus re-ranked to the first position, as illustrated in Figure~\ref{fig:framework-overview}.
Unlike existing methods~\cite{imagescope,cotmr} in which the MLLM is used solely for evaluation, our approach leverages the MLLM to perform both evaluation and ranking within one stage. This one-stage process not only simplifies the overall pipeline but also enables the model to make more globally consistent ranking decisions based on a holistic understanding of the evaluation results. 
This final ranking step is formally represented as:  
\[
\mathcal{R}_{\text{Top-K}} = \text{Rank}(\mathcal{V})
\]

\section{Experiments}

\subsection{Datasets and experimental settings}
\textbf{Datasets}
We evaluate our proposed CoTRR on five datasets across three tasks: TIR, CIR, and Chat-IR. For TIR, we use the widely adopted Flickr30K~\cite{flickr30k} and MSCOCO~\cite{mscoco} datasets, both providing multiple human-annotated captions per image. For CIR, we adopt CIRR~\cite{cirr} and CIRCO~\cite{circo}; CIRR is the first dataset designed for CIR despite some false negatives~\cite{baldrati2023zero}, which CIRCO mitigates by offering multiple annotated ground truths per query. For Chat-IR, we use the human-annotated VisDial dataset~\cite{visdial} and evaluate multi-round retrieval with Hits@k. Following original benchmarks, Recall@k (R@k) is used for Flickr30K, MSCOCO, and CIRR, while CIRCO is evaluated with mean average precision at k (mAP@k). Additionally, $\text{Recall}_{\text{Subset}}\text{@k}$ ($\text{R}_{\text{Sub}}\text{@k}$) is reported for CIRR to measure the retrieval performance within a subset. 

\noindent \textbf{Baselines}
We compare our CoTRR with a variety of strong baselines. As our CoTRR is training-free, we focus on training-free baselines for fair comparison. 
(1) For CIR, we compare against CIReVL~\cite{CIReVL}, OSrCIR~\cite{osrcir}, and ImageScope~\cite{imagescope}. (2) For TIR, we compare CoTRR with OpenCLIP~\cite{openClip} and ImageScope~\cite{imagescope} to highlight its performance improvements.
(3)For Chat-IR, We assess the effectiveness of CoTRR by comparing it with OpenCLIP~\cite{openClip}, PlugIR~\cite{plugir}, and ImageScope~\cite{imagescope}.

\noindent \textbf{Implementation Details} The default VLM used in CoTRR is Gemini~2.5-Pro~\cite{gemini}. We also conduct ablation studies using {Gemini~2.5-Flash}~\cite{gemini}, {GPT-4o}~\cite{gpt4o}, and {Qwen-VL-Max}~\cite{qwen-vl}. 
{The temperature is set to 0 for all API calls, while all other parameters are kept at their default values.}
For the CLIP backbone, we employ two variants from OpenCLIP~\cite{openClip}: {ViT-B/32} and {ViT-L/14}. 
All experiments are conducted on a single NVIDIA RTX 3090 GPU.
{The number of top-K candidates for re-rank is set to 20 for Flickr30k, MSCOCO and VisDial, to 15 and 70 for CIRR and CIRCO respectively, to 3 for CIRR subset}

\begin{figure*}[t]
    \centering
    \includegraphics[width=1.0\textwidth]{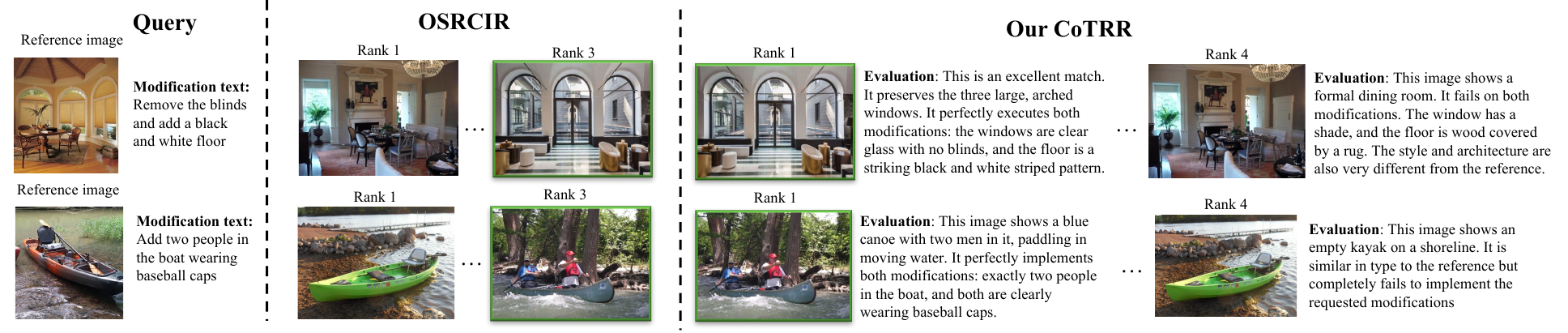}
    \caption{A qualitative comparison between OSRCIR (our baseline) and CoTRR demonstrates that our method effectively re-ranks the ground-truth image to a correct position, addressing the failure in OSRCIR.}
    \label{fig:exmaple}
\end{figure*}

\vspace{-8pt}
\subsection{Performance Evaluation}
\noindent \textbf{Composed image retrieval} Table~\ref{tab:CIRResult} shows quantitative comparison results  on CIRR and CIRCO test set. As shown,  our CoTRR achieves the state-of-the-art performance compared to the baselines on both CIRR and CIRCO datasets using different CLIP-based ViT variants as backbone.
With {ViT-B/32}, CoTRR surpasses ImageScope with absolute improvements of 12.41\% in R@1 and 6.53\% in R@5 on the CIRR dataset, and 16.10\% in mAP@5 and 15.49\% in mAP@10 on the CIRCO dataset. 
This demonstrates the effectiveness of CoTRR in enhancing retrieval performance via re-ranking, especially with significant gains in R@1. 

\begin{figure}[htbp]
    \centering
    \includegraphics[width=\columnwidth]{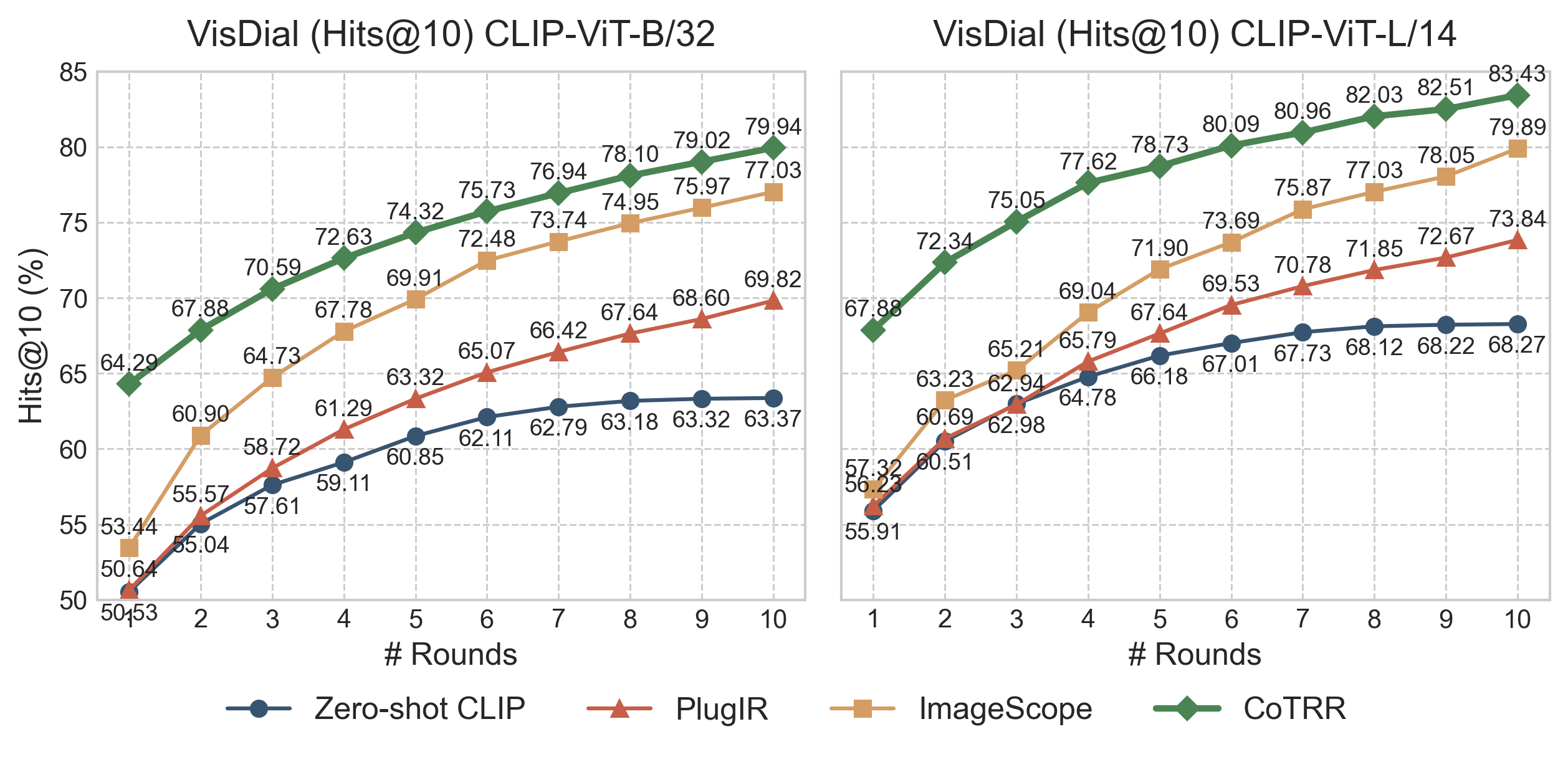}
    \caption{Performance of Chat-IR on VisDial \cite{visdial} compared with Zero-shot CLIP \cite{openClip}, PlugIR \cite{plugir}, and ImageScope \cite{imagescope}.}
    \label{fig:my_line_chart}
\end{figure}

\noindent \textbf{Text-to-image retrieval and chat-based image retrieval} 
 Table \ref{tab:TIRResults} presents the performance of CoTRR compared to the original CLIP and ImageScope on the Flickr30K and MSCOCO datasets. 
 It can be observed that our CoTRR outperforms both CLIP and ImageScope across the R@1, R@5, and R@10 metrics. This shows the effectiveness of our CoTRR on text-to-image retrieval. Figure~\ref{fig:my_line_chart} presents the comparative results for Chat-IR among our CoTRR, OpenCLIP~\cite{openClip}, PlugIR~\cite{plugir}, and ImageScope~\cite{imagescope}. It can be observed that CoTRR consistently outperforms the other methods across multiple dialogue rounds, using different CLIP-based ViT variants for retrieval. For instance, compared to ImageScope~\cite{imagescope}, our method achieves a 10.56\% improvement in the first round and a 6.83\% improvement in the fifth round when  using
{ViT-L/14}.

\subsection{Ablation Study}
We first investigate the contribution of each component in our CoTRR, including the List-wise Ranking (R), Deconstruction (D), and Image Evaluation (E) modules. We then analyze the influence of the different MLLMs on retrieval performance. 


\noindent \textbf{Effect of each component in CoTRR} As shown in Table~\ref{tab:Ablation_CIRR_CoT}, applying the ranking module alone (model '2')  brings a substantial improvement over the baseline (model '1'), raising R@1 from 32.63 to 47.11. 
This highlights the effectiveness and importance of the direct re-ranking process in reassessing the initial candidates based on how well they satisfy the user's input.
Adding the deconstruction module (model '3') further enhances performance to 50.07 in terms of R@1, indicating that decomposing complex queries into structured sub-components benefits the comparison and ranking process. Similarly, combining R and E (model '4') also improves R@5 from 70.15 to 70.63, though slightly less than R+D. 
The full model with all three components (model '5') achieves the best performance, reaching 50.84 in R@1 and 72.80 in R@5, demonstrating that the three modules work complementarily to enhance retrieval performance. 
 Figure~\ref{fig:exmaple} shows a qualitative comparison between OSRCIR (our baseline) and our CoTRR. It can be seen that incorrect candidate rankings in OSRCIR are corrected by our CoTRR. 

\noindent \textbf{Effect of different MLLMs} We further examine the effect of different MLLMs on CoTRR's performance by keeping the prompts fixed and varying only the underlying MLLM. As shown in Table~\ref{tab:Ablation_CIRR_CoT}, Gemini 2.5 Pro yields the best performance, 50.84 in terms of R@1. Qwen-VL-Max also performs competitively with 49.52, slightly outperforming Gemini 2.5 Flash (48.75) and GPT-4o (47.83). These results suggest that  our CoTRR is  robust across different MLLMs, and larger or more instruction-aligned MLLMs tend to provide better grounding and evaluation capabilities, which are critical for our re-ranking process.






\begin{table}[]
\centering
\caption{Ablation study on CIRR evaluating the contribution of CoTRR components and the effect of different MLLMs on retrieval performance.}
\label{tab:Ablation_CIRR_CoT}
\begin{tabular}{p{4cm}ccc}
\hline
\multirow{2}{*}{Method} & \multicolumn{3}{c}{CIRR} \\
\cline{2-4}
& R@1 & R@5 & R@10 \\
\hline
\multicolumn{4}{l}{\textbf{Significance of each component in CoTRR}} \\
1. Baseline &  32.63 & 61.23& 74.02\\ 
2. with Only R  & 47.11 & 70.15 & 77.06 \\
3. with R+D & 50.07 & 71.25 & 77.25 \\
4. with R+E & 47.11 & 70.63 & 77.16 \\
5. with R+D+E  & 50.84 & 72.80 & 77.83 \\
\midrule
\multicolumn{4}{l}{\textbf{Performance of our full model with different MLLMs}} \\
\midrule
Gemini2.5 pro \cite{gemini} & 50.84 & 72.80 & 77.83 \\
6. Gemini2.5 flash \cite{gemini} & 48.75 & 71.42 & 77.35 \\
7. Qwen-VL-Max \cite{qwen-vl} & 49.52 & 71.59 & 77.54 \\
8. GPT-4o \cite{gpt4o} & 47.83 & 70.01 & 76.70 \\
\hline
\end{tabular}
\end{table}


\section{Conclusion}
In this paper, we introduced CoTRR, a novel chain-of-thought re-ranking method to enhance image ranking across various retrieval tasks. Specifically, CoTRR employs a query deconstruction prompt to thoroughly understand a user's intention and reformat the input into a standardized structure, an intuitive image evaluation paradigm for detailed candidate assessment, and a listwise ranking prompt to enable more effective comparison between the user’s input and candidate image. Experimental results on five datasets demonstrate that our CoTRR achieves state-of-the-art performance on three retrieval tasks. Ablation study also demonstrates the effectiveness of CoTRR.

\vfill\pagebreak



\bibliographystyle{IEEEbib}

\bibliography{strings,refs }

\begin{thebibliography}{10}

\bibitem{datta2008image}
Ritendra Datta, Dhiraj Joshi, Jia Li, and James~Z Wang,
\newblock ``Image retrieval: Ideas, influences, and trends of the new age,''
\newblock {\em ACM Computing Surveys (Csur)}, vol. 40, no. 2, pp. 1--60, 2008.

\bibitem{Dialog-based}
Xiaoxiao Guo, Hui Wu, Yu~Cheng, Steven Rennie, Gerald Tesauro, and Rogerio Feris,
\newblock ``Dialog-based interactive image retrieval,''
\newblock {\em Advances in neural information processing systems}, vol. 31, 2018.

\bibitem{chatir}
Matan Levy, Rami Ben-Ari, Nir Darshan, and Dani Lischinski,
\newblock ``Chatting makes perfect: Chat-based image retrieval,''
\newblock {\em Advances in Neural Information Processing Systems}, vol. 36, pp. 61437--61449, 2023.

\bibitem{ma2025multi}
Zehong Ma, Hao Chen, Wei Zeng, Limin Su, and Shiliang Zhang,
\newblock ``Multi-modal reference learning for fine-grained text-to-image retrieval,''
\newblock {\em IEEE Transactions on Multimedia}, 2025.

\bibitem{wen2023target}
Haokun Wen, Xian Zhang, Xuemeng Song, Yinwei Wei, and Liqiang Nie,
\newblock ``Target-guided composed image retrieval,''
\newblock in {\em Proceedings of the 31st ACM international conference on multimedia}, 2023, pp. 915--923.

\bibitem{plugir}
Saehyung Lee, Sangwon Yu, Junsung Park, Jihun Yi, and Sungroh Yoon,
\newblock ``Interactive text-to-image retrieval with large language models: A plug-and-play approach,''
\newblock {\em arXiv preprint arXiv:2406.03411}, 2024.

\bibitem{osrcir}
Yuanmin Tang, Jue Zhang, Xiaoting Qin, Jing Yu, Gaopeng Gou, Gang Xiong, Qingwei Lin, Saravan Rajmohan, Dongmei Zhang, and Qi~Wu,
\newblock ``Reason-before-retrieve: One-stage reflective chain-of-thoughts for training-free zero-shot composed image retrieval,''
\newblock in {\em Proceedings of the Computer Vision and Pattern Recognition Conference}, 2025, pp. 14400--14410.

\bibitem{cotmr}
Zelong Sun, Dong Jing, and Zhiwu Lu,
\newblock ``Cotmr: Chain-of-thought multi-scale reasoning for training-free zero-shot composed image retrieval,''
\newblock {\em arXiv preprint arXiv:2502.20826}, 2025.

\bibitem{imagescope}
Pengfei Luo, Jingbo Zhou, Tong Xu, Yuan Xia, Linli Xu, and Enhong Chen,
\newblock ``Imagescope: Unifying language-guided image retrieval via large multimodal model collective reasoning,''
\newblock in {\em Proceedings of the ACM on Web Conference 2025}, 2025, pp. 1666--1682.

\bibitem{wei2022chain}
Jason Wei, Xuezhi Wang, Dale Schuurmans, Maarten Bosma, Fei Xia, Ed~Chi, Quoc~V Le, Denny Zhou, et~al.,
\newblock ``Chain-of-thought prompting elicits reasoning in large language models,''
\newblock {\em Advances in neural information processing systems}, vol. 35, pp. 24824--24837, 2022.

\bibitem{zhang2023multimodal}
Zhuosheng Zhang, Aston Zhang, Mu~Li, Hai Zhao, George Karypis, and Alex Smola,
\newblock ``Multimodal chain-of-thought reasoning in language models,''
\newblock {\em arXiv preprint arXiv:2302.00923}, 2023.

\bibitem{clip1}
Alec Radford, Jong~Wook Kim, Chris Hallacy, Aditya Ramesh, Gabriel Goh, Sandhini Agarwal, Girish Sastry, Amanda Askell, Pamela Mishkin, Jack Clark, et~al.,
\newblock ``Learning transferable visual models from natural language supervision,''
\newblock in {\em International conference on machine learning}. PmLR, 2021, pp. 8748--8763.

\bibitem{CIReVL}
Shyamgopal Karthik, Karsten Roth, Massimiliano Mancini, and Zeynep Akata,
\newblock ``Vision-by-language for training-free compositional image retrieval,''
\newblock {\em arXiv preprint arXiv:2310.09291}, 2023.

\bibitem{flickr30k}
Peter Young, Alice Lai, Micah Hodosh, and Julia Hockenmaier,
\newblock ``From image descriptions to visual denotations: New similarity metrics for semantic inference over event descriptions,''
\newblock {\em Transactions of the association for computational linguistics}, vol. 2, pp. 67--78, 2014.

\bibitem{mscoco}
Tsung-Yi Lin, Michael Maire, Serge Belongie, James Hays, Pietro Perona, Deva Ramanan, Piotr Doll{\'a}r, and C~Lawrence Zitnick,
\newblock ``Microsoft coco: Common objects in context,''
\newblock in {\em European conference on computer vision}. Springer, 2014, pp. 740--755.

\bibitem{cirr}
Zheyuan Liu, Cristian Rodriguez-Opazo, Damien Teney, and Stephen Gould,
\newblock ``Image retrieval on real-life images with pre-trained vision-and-language models,''
\newblock in {\em Proceedings of the IEEE/CVF international conference on computer vision}, 2021, pp. 2125--2134.

\bibitem{circo}
Alberto Baldrati, Lorenzo Agnolucci, Marco Bertini, and Alberto Del~Bimbo,
\newblock ``Zero-shot composed image retrieval with textual inversion,''
\newblock in {\em Proceedings of the IEEE/CVF International Conference on Computer Vision}, 2023, pp. 15338--15347.

\bibitem{baldrati2023zero}
Alberto Baldrati, Lorenzo Agnolucci, Marco Bertini, and Alberto Del~Bimbo,
\newblock ``Zero-shot composed image retrieval with textual inversion,''
\newblock in {\em Proceedings of the IEEE/CVF International Conference on Computer Vision}, 2023, pp. 15338--15347.

\bibitem{visdial}
Abhishek Das, Satwik Kottur, Khushi Gupta, Avi Singh, Deshraj Yadav, Jos{\'e}~MF Moura, Devi Parikh, and Dhruv Batra,
\newblock ``Visual dialog,''
\newblock in {\em Proceedings of the IEEE conference on computer vision and pattern recognition}, 2017, pp. 326--335.

\bibitem{openClip}
Chao Jia, Yinfei Yang, Ye~Xia, Yi-Ting Chen, Zarana Parekh, Hieu Pham, Quoc Le, Yun-Hsuan Sung, Zhen Li, and Tom Duerig,
\newblock ``Scaling up visual and vision-language representation learning with noisy text supervision,''
\newblock in {\em International conference on machine learning}. PMLR, 2021, pp. 4904--4916.

\bibitem{gemini}
Gheorghe Comanici, Eric Bieber, Mike Schaekermann, Ice Pasupat, Noveen Sachdeva, Inderjit Dhillon, Marcel Blistein, Ori Ram, Dan Zhang, Evan Rosen, et~al.,
\newblock ``Gemini 2.5: Pushing the frontier with advanced reasoning, multimodality, long context, and next generation agentic capabilities,''
\newblock {\em arXiv preprint arXiv:2507.06261}, 2025.

\bibitem{gpt4o}
Aaron Hurst, Adam Lerer, Adam~P Goucher, Adam Perelman, Aditya Ramesh, Aidan Clark, AJ~Ostrow, Akila Welihinda, Alan Hayes, Alec Radford, et~al.,
\newblock ``Gpt-4o system card,''
\newblock {\em arXiv preprint arXiv:2410.21276}, 2024.

\bibitem{qwen-vl}
Jinze Bai, Shuai Bai, Shusheng Yang, Shijie Wang, Sinan Tan, Peng Wang, Junyang Lin, Chang Zhou, and Jingren Zhou,
\newblock ``Qwen-vl: A frontier large vision-language model with versatile abilities,''
\newblock {\em arXiv preprint arXiv:2308.12966}, vol. 1, no. 2, pp. 3, 2023.

\end{thebibliography}
\vfill
\begin{center}
\footnotesize
This work has been submitted to the IEEE for possible publication. Copyright may be transferred without notice, after which this version may no longer be accessible.
\end{center}

\end{document}